# Developing an Explainable Artificial Intelligent (XAI) Model for Predicting Pile Driving Vibrations in Bangkok's Subsoil

Sompote Youwai*
Anuwat Pamungmoon

Artificial Intelligent Research Group
Department of Civil Engineering King Mongkut's University of Technology Thonburi
*Corresponding Author: sompote.you@kmutt.ac.th

**Abstract**
This study presents an explainable artificial intelligent (XAI) model for predicting pile driving vibrations in Bangkok's soft clay subsoil. A deep neural network was developed using a dataset of 1,018 real-world pile driving measurements, encompassing variations in pile dimensions, hammer characteristics, sensor locations, and vibration measurement axes. The model achieved a mean absolute error (MAE) of 0.276, outperforming traditional empirical methods and other machine learning approaches such as XGBoost and CatBoost. SHapley Additive exPlanations (SHAP) analysis was employed to interpret the model's predictions, revealing complex relationships between input features and peak particle velocity (PPV). Distance from the pile driving location emerged as the most influential factor, followed by hammer weight and pile size. Non-linear relationships and threshold effects were observed, providing new insights into vibration propagation in soft clay. A web-based application was developed to facilitate adoption by practicing engineers, bridging the gap between advanced machine learning techniques and practical engineering applications. This research contributes to the field of geotechnical engineering by offering a more accurate and nuanced approach to predicting pile driving vibrations, with implications for optimizing construction practices and mitigating environmental impacts in urban areas. The model and its source code are publicly available, promoting transparency and reproducibility in geotechnical research.

# 1. Introduction

In urban development, pile driving is a critical construction technique that reinforces structural stability and integrity. Nevertheless, it induces ground vibrations with potential adverse impacts on adjacent environments and infrastructures. The vibrations from pile driving may inflict damage on nearby structures or buildings. Peak Particle velocity (PPV) serves as a key parameter reflecting the extent of structural damage attributable to pile driving. PPV's magnitude is contingent upon variables including proximity, pile dimensions, and hammer attributes. Consequently, evaluating these vibrations is essential for damage prevention and the safeguarding of public welfare.

Previous research endeavors have aimed to quantify the vibrational impacts stemming from pile driving operations. These investigations have yielded predictive models for PPV and distance proximal to pile driving, utilizing empirical formulations (Attewell and Farmer 1973; Wiss 1981; Achmus et al. 2010; Massarsch and Fellenius 2015; Grizi et al. 2016; Weng et al. 2020). However, notable discrepancies have been documented between empirical forecasts and actual in-situ test results. In a recent development, finite element analysis has been applied to anticipate ground vibrations (Wang and Zhu 2023), though the predictive outcomes were corroborated with a limited dataset. Despite the comprehensive nature of finite element analysis, it is characterized by its extensive duration and significant financial implications. Furthermore, critical issues have emerged, as prior studies were predicated on sandy soil substrates, which exhibit lower damping characteristics in comparison to clay profiles. The latter may experience augmented vibrational effects from pile driving, a condition prevalent in the Bangkok region's typical soil composition. It can be hypothesized that current methodologies might overestimate the peak particle velocity (PPV). Consequently, there exists an urgent need for a more expedient and precise method to accurately predict vibrations induced by pile driving activities.

Explainable Artificial Intelligence (XAI) represents a critical domain within AI research, focusing on enhancing the transparency and interpretability of AI decision-making processes. The proliferation of AI technologies across diverse sectors, including healthcare (Lai 2024) and finance (Chen et al. 2023), underscores the imperative for comprehensible AI systems. Many contemporary AI models, particularly those employing deep learning algorithms, operate as opaque "black boxes," presenting significant challenges in understanding their internal mechanisms. The primary objective of XAI is to elucidate the reasoning underlying AI decisions, thereby augmenting user trust and accountability. By providing comprehensive explanations, XAI facilitates user comprehension of the factors influencing outcomes, thus promoting perceptions of fairness and reliability. In the geotechnical research domain, XAI methodologies have been applied to elucidate model results in areas such as slope stability(Abdollahi et al. 2024) analysis, liquefaction prediction (Hsiao et al. 2024), and tunneling assessments (Liu et al. 2024) . XAI techniques can explicate instances where models fail to accurately classify actual values, highlighting the relative effectiveness of various parameters. Furthermore, XAI enables the interpretation of model outputs at different certainty levels, elucidating the contributions of individual parameters to both local and global predictions. Trained models, when analyzed through XAI frameworks, can yield insights into the influence of specific features on model characteristics. In the geotechnical engineering field, the explication of black-box models through XAI methodologies can assist geotechnical engineers who lack expertise in deep learning architectures to comprehend trained model characteristics and develop confidence in applying these models to real-world scenarios.

This study aims to develop and validate a deep neural network model for predicting pile driving vibrations in Bangkok subsoil using a comprehensive dataset of real-world measurements, while analyzing the influence of various parameters such as pile dimensions, hammer characteristics, sensor locations, and vibration measurement axes. The research objectives include implementing and training the neural network model using the PyTorch library (Paszke et al. 2019), conducting an in-depth interpretability analysis using SHapley Additive exPlanations (SHAP) (Wei et al. 2021; Li et al. 2024) to elucidate the model's decision-making process, and investigating the complex behavior of pile driving vibrations in Bangkok subsoil through quantitative assessment of SHAP values. Furthermore, the study seeks to compare the performance of the developed deep learning model against traditional empirical methods, design and develop a user-friendly web application for practical use by geotechnical engineers, and validate both the model and web application through case studies on ongoing construction projects in Bangkok. Ultimately, this research aims to demonstrate the model's practical applicability and its potential for improving vibration control strategies in urban environments, thereby advancing the field of geotechnical engineering and contributing to more efficient and safer construction practices in areas with similar soil conditions.

Key findings include:

- The deep neural network model achieved a mean absolute error of 0.276 mm/s, demonstrating its efficacy for practical vibration estimation applications.
- A comparative analysis between the proposed model and existing empirical formulas revealed enhanced accuracy of the neural network approach.
- This study represents the first investigation of pile driving-induced vibrations specifically in Soft Bangkok Clay, addressing a significant gap in geotechnical literature.
- Explainable AI (XAI) techniques were employed to interpret model results under diverse conditions.
- A web-based application was developed to facilitate adoption by practicing engineers, enabling convenient estimation of pile vibrations during driving operations.
- The model is publicly available on the authors' GitHub repository and can be implemented using Python, enhancing accessibility for researchers and practitioners.

## 2. Data characteristics

A dataset of 1,034 observations was collected from an active construction site in Bangkok's soft soil area. The site featured characteristic Bangkok subsoil conditions, predominantly consisting of very soft clay in the upper layers. Specifically, very soft clay extended from the surface to approximately 18 m depth, followed by layers of medium and stiff clay. The first sand layer was identified at approximately 24-30 m depth ((Mairaing and Amonkul 2010; Boonyatee et al. 2015). Figure 1 depicts the typical experimental setup, including piles of various diameters and lengths. The diagram illustrates diverse driving criteria, sensor locations, and measurement orientations. Dataset characteristics are also summarized in Figure 2. Hammer weight ranged from 3 to 12.5 tons, while pile length varied between 10 and 32 meters. Hammer drop height spanned 0.3 to 0.9 meters, with a modal distribution between 0.3 and 0.4 m. Pile diameters ranged from 0.25 to 0.8 meters. The primary variable of interest, distance from the pile driving location, was measured at intervals of 3 to 80 meters. Sensor placement was categorized into three locations: ground (1), footing (2), and building (3), with the majority positioned on buildings. Sensor orientation, referred to as the trigger, was designated as longitudinal (1), transverse (2), or vertical (3), with a predominance of longitudinal orientations relative to the pile driving location. The model's output variable, peak particle velocity, exhibited substantial variation, ranging from 8 mm/s to approximately 0.127 mm/s.

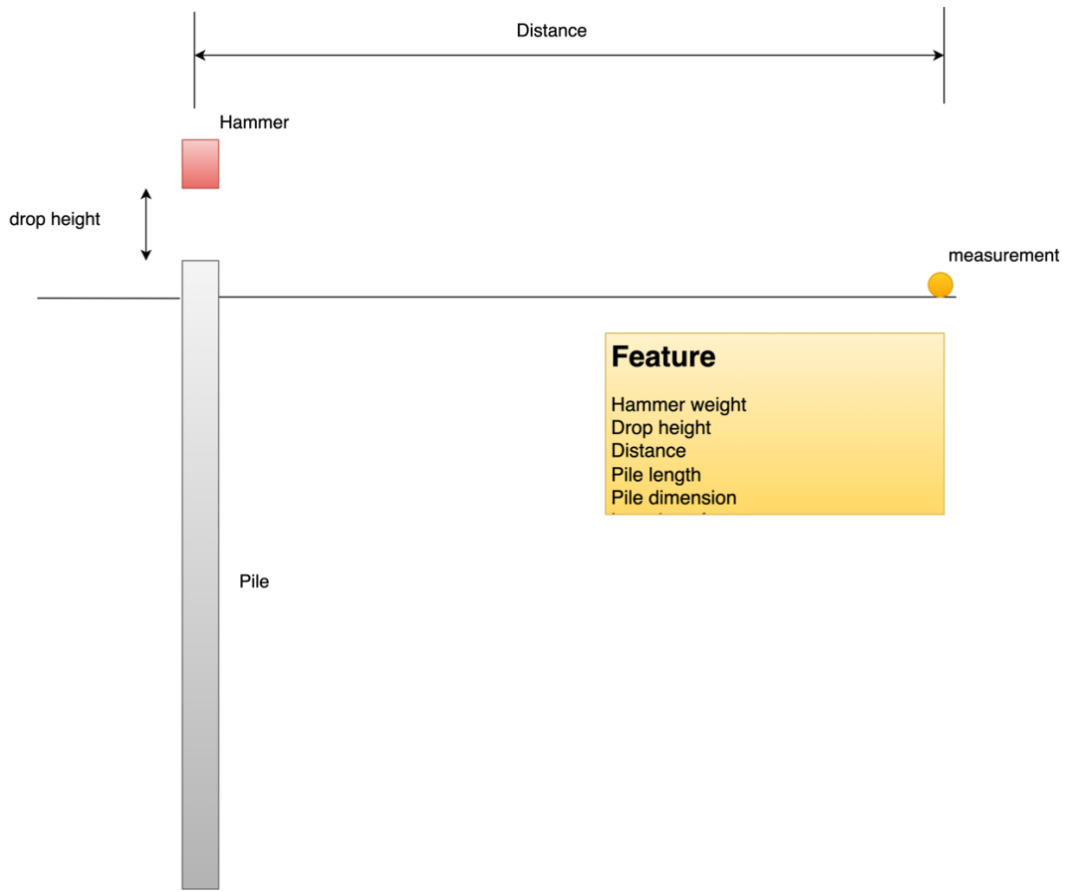

**Figure 1** Schematic diagram for vibration prediction

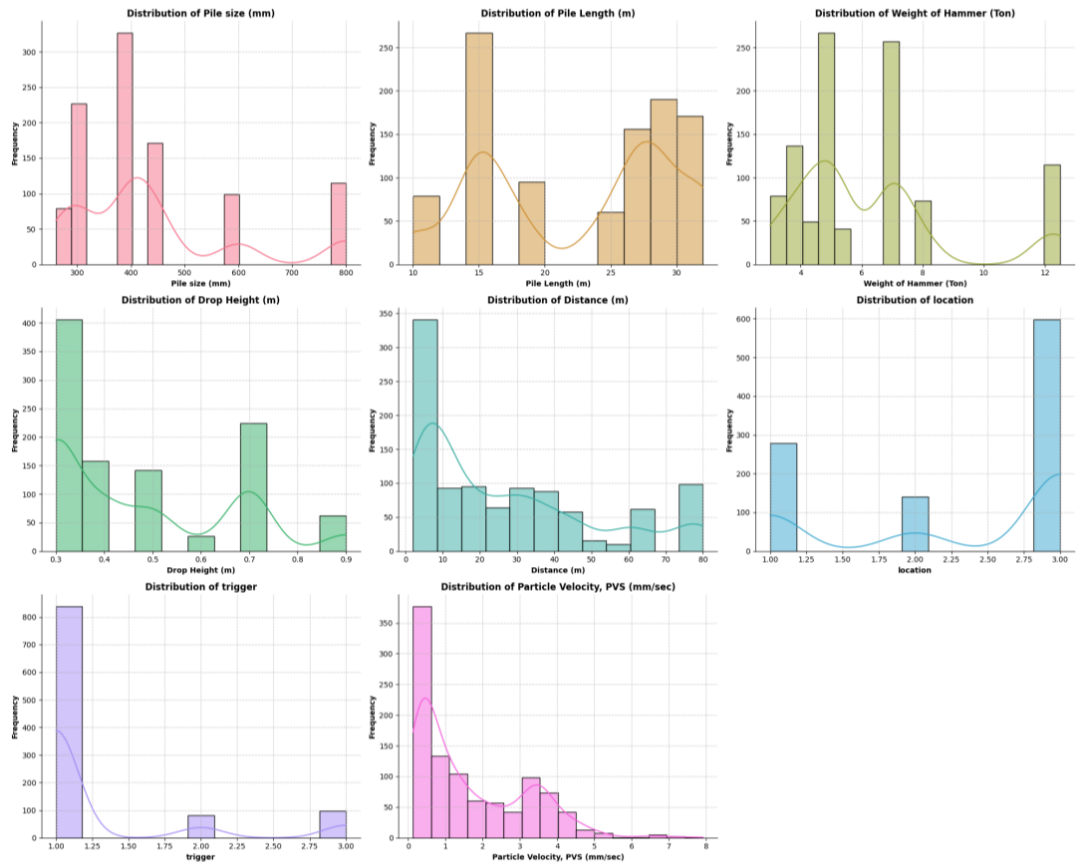

Figure 2 The distribution of feature for model

The mutual information (Kraskov et al. 2004) of each influencing factor on the maximum particle velocity is depicted in Figure 3. Mutual information measures how much the knowledge of one variable reduces the uncertainty of the other. It is a powerful tool because it can detect any kind of relationship between variables, not just linear ones like correlation does. The proximity of the pile driving location to the measurement point exhibited a significant influence. Subsequently, the mass of the hammer and the length of the pile were identified as influential factors affecting the vibration of the pile. These two factors were the primary contributors to the generation of vibrations, stemming from the energy of the applied vibrations. An increased pile length could potentially extend the source of vibration to the ground surface. Following these, the dimensions of the pile and the height from which the hammer is dropped were also influential. Conversely, the location of the sensor and the orientation of the vibration measurement (trigger) were determined to have a minimal influence.

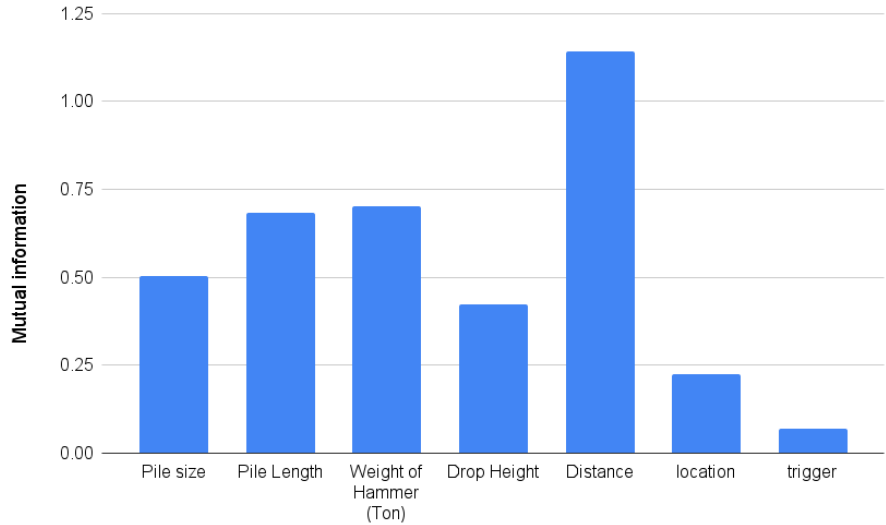

Figure 3 The mutual information of the feature to the peak particle velocity

3. Proposed Model

The model was developed using the PyTorch (Paszke et al. 2019) due to its flexibility in custom architecture design and efficient inference capabilities. PyTorch's dynamic computational graph and native support for GPU acceleration facilitated rapid prototyping and optimization of the model. The implementation was carried out in Python, leveraging its extensive ecosystem of scientific computing libraries. The architecture of the model is based on a multilayer perceptron (MLP), a class of feedforward artificial neural networks. Specifically, the model employs a deep MLP structure with multiple hidden layers, allowing it to capture complex, non-linear relationships in the pile driving vibration data. The input layer of the model is structured upon an architecture that incorporates a linear transformation, as delineated by the following equation:

$$y = xA^T + b \qquad (1)$$

As shown in Figure 4, the model's architecture comprises a series of linear transformations arranged in a diamond-shaped structure. This design expands in the middle layers to enhance feature extraction capabilities, allowing the network to capture and process a wider range of features in its intermediate representations. The model begins with 7 perceptrons, matching the number of input features. It then expands to 100, 200, and 20 perceptrons in subsequent layers, forming the widest part of the diamond. To improve generalization performance, a dropout layer with a rate of 0.1 is applied after this expansion. The architecture then narrows to 5 perceptrons, followed by a single output perceptron predicting the peak particle velocity.

Activation functions are applied between the linear layers to introduce non-linear characteristics to the model, enabling it to learn complex patterns in the data. Rectified Linear Units (ReLU) are primarily used throughout the network due to their computational efficiency and effectiveness in mitigating the vanishing gradient problem. The ReLU function can be expressed as:

$$sigmoid(x) = \frac{1}{1+e^{-x}} \qquad (2)$$

However, the final layer employs a Sigmoid activation function to normalize the output, making it suitable for this regression problem. The Sigmoid function, which maps the output to a range between 0 and 1, is defined as:

$$ReLU(x) = \max(0, x) \qquad (3)$$

To regularize the model and prevent overfitting, a dropout layer is integrated into the architecture. This layer functions by randomly deactivating approximately 10% of the neurons during training, thereby improving the model's generalization capabilities.

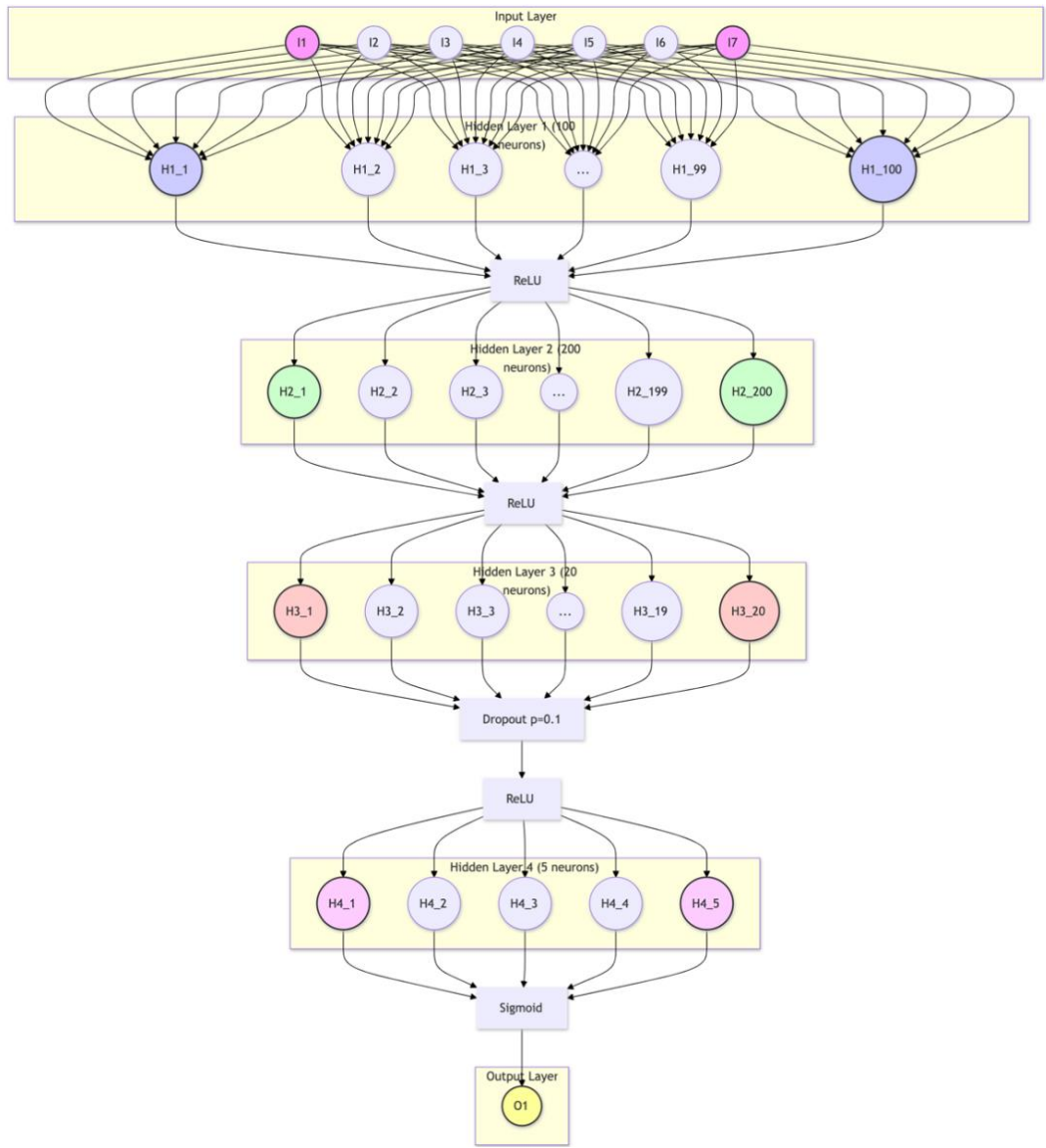

**Figure. 4 The architecture of deep neural network model**

The input tensor was configured with dimensions (size, 7), where it represents the number of data instances processed concurrently during training. The tensor comprises seven features, as delineated in Table 1. The 'location of sensor' and 'direction of sensor' attributes are categorized as categorical features, encoded numerically as 1, 2, or 3. The remaining features are treated as numerical types. Standardization of all feature types was performed utilizing a StandardScaler, as per the equation below:

$$z = \frac{x_i - \mu}{\sigma} \quad (4)$$

| Feature | Detail |
| --- | --- |

| | | |
|---|---|---|
| Pile size | mm | Table 1 The detail of the features |
| Pile length | m | |
| Weight of hammer | ton | |
| Drop height | m | |
| Distance | m | |
| Location of sensor | 1: on ground 2: on footing 3: on building | |
| Direction of sensor | 1: longitudinal 2: transverse 3: vertical | |

## 4. Experiment

The neural network underwent training to refine the weight matrix, with the objective of diminishing the loss function. The dataset was segregated into three distinct subsets: training, validation, and testing, adhering to a distribution ratio of 80:10:10, respectively. The validation subset played a pivotal role during the training phase, facilitating the fine-tuning of hyperparameters. In contrast, the test subset, composed of data not exposed to the model during training, was instrumental in ascertaining the model's generalization capacity. It is of paramount importance to ensure that the model does not succumb to overfitting with respect to the training data. Throughout the training process, data was amalgamated into batches of 50, and the gradient within each batch was consolidated to update the weight matrix. The rationale behind employing batch gradient combination, as opposed to individual data point training, lies in its ability to mitigate fluctuations or noise in the weight trajectory during training. The update of the weight matrix within the deep learning architecture was executed through the backpropagation of the gradient. The loss of the trained model was mean square error due to the problem is regression analysis as follows:

$$\mathcal{L} = \frac{1}{n} \sum_{i=1}^{n}(y_i - \hat{y}_i)^2 \qquad (5)$$

The learning rate was dynamically modulated utilizing the Adam optimization algorithm (Kingma and Ba 2014), which adjusts the learning rate for each weight predicated on the exponential moving averages of the gradients and their squared counterparts. This methodology is acclaimed for its operational efficiency, minimal memory footprint, and compatibility with voluminous datasets and non-stationary objectives. The performance matrix of the model was mean absolute error (MAE) as follows:

$$\text{MAE} = \frac{1}{n}\sum_{i=1}^{n} | y_i - \hat{y}_i | \qquad (6)$$

The training results are depicted in Figure 5. Notably, the model exhibited favorable behavior with respect to overfitting. As the training progressed beyond 200 epochs, the loss for both the training and validation datasets converged to a stable level. Subsequently, the trend became nearly constant, indicating an appropriate point to halt training. In this research, the stopping point was deliberately set at 500 epochs. The best performance metrics from the validation dataset were saved as the model's optimal training behavior. This best-performing model was then used to evaluate performance on the test dataset. The results from the trained model validation with test data yielded a Mean Absolute Error (MAE) of 0.276. A comparison with other model types will be presented and discussed in the next part of this study (Ablation study). The trained model was then employed to predict the overall data, including test, train, and validation sets, as shown in Figure 6. The results show moderate scatter, particularly for data points with high peak particle velocity values or for sensor locations near the vibration source. This scatter is a common phenomenon in field test data collected from various situations, such as different soil profiles. However, the variation remains within an acceptable range, especially for peak particle velocity values lower than 3 mm/s.

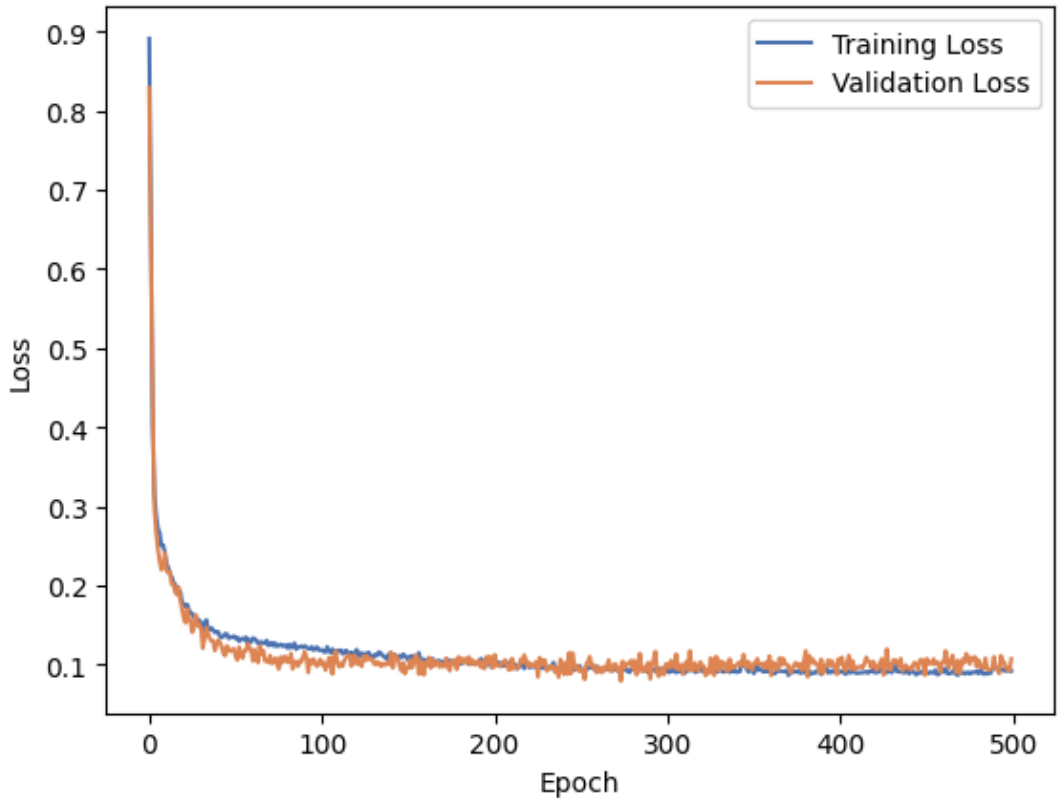

Figure 5 The loss of model during training

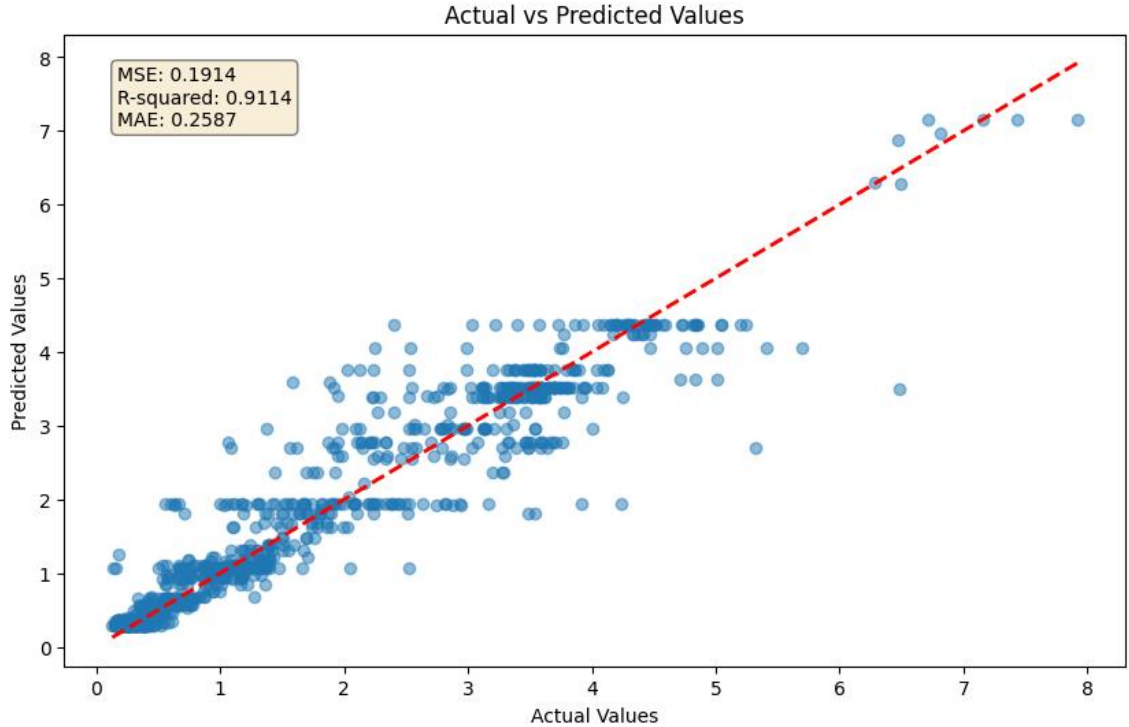

**Figure 6** The comparison between the predicted value and actual value

### 4.1 Ablation study

An ablation study was conducted to optimize the architecture of the deep neural network model for predicting pile driving vibrations, aiming to identify the most effective combination of model components, focusing on perceptron size and activation function types. The study, detailed in Table 2, involved training various model configurations with a batch size of 50 for 400 epochs, using Mean Absolute Error (MAE) as the performance metric. Four activation functions were tested: Sigmoid, Tanh, ReLU, and LeakyReLU. Results revealed that ReLU consistently yielded the lowest MAE, indicating its superior suitability for capturing patterns in pile driving vibration data. The optimized model architecture incorporating ReLU outperformed both simpler (fewer layers) and more complex (higher number of perceptrons) alternatives, suggesting an optimal balance between model complexity and generalization ability. The performance of the optimized model was also compared with XGBoost and CatBoost models, two prominent gradient boosting frameworks, with the proposed deep neural network model demonstrating superior performance. This superiority can be attributed to several factors: deep learning models excel at capturing intricate, non-linear relationships between features prevalent in geotechnical data; their architecture is well-suited to capture spatial correlations across different scales; they create a continuous representation more appropriate for vibration data; and they are inherently designed to handle and benefit from high-dimensional data complexity. This rigorous optimization process enhances the model's robustness and reliability, demonstrating a scientific approach to developing an efficient and accurate predictive tool for pile driving vibrations. The findings underscore the importance of careful model architecture selection in achieving optimal performance for specific geotechnical applications and highlight the potential advantages of deep learning approaches in this domain.

Table 2 The ablation study of the model

| Model | MAE |
|---|---|
| XGboost (Chen and Guestrin 2016) | 0.315 |
| Catboost (Prokhorenkova et al. 2018) | 0.332 |
| (7,200,1000,2000,200,20,5,1):ReLU | 0.289 |
| (7,50,100,20,5,1):ReLU | 0.283 |
| (7,100,200,20,5,1): Sigmoid | 0.432 |
| (7,100,200,20,5,1):Tanh | 0.279 |
| (7,100,200,20,5,1):LeakyReLU | 0.852 |
| **Propose model: ReLU (7,100,200,20,5,1)** | **0.276** |

bold denoted the best performance

## 5. Explainable neural network

SHAP (SHapley Additive exPlanations), developed by (Lundberg and Lee 2017) and further explored by (Li et al. 2024), is a sophisticated method widely used to interpret individual predictions made by machine learning models. It quantifies the contribution of each feature to the prediction, providing valuable insights into model behavior. Rooted in Shapley values from cooperative game theory, SHAP offers a systematic approach to distribute the total prediction value among features based on their individual contributions. In this context, the "players" are the model's features, and the "value" is the model's prediction. SHAP primarily focuses on understanding the impact of each attribute on the prediction for a specific data point. This method has gained significant traction in explaining machine learning models across various domains. For instance, in a linear model, SHAP can elucidate how each feature contributes to the final prediction, providing a clear breakdown of feature importance and direction of influence. This approach is particularly valuable in complex models where feature interactions and non-linear relationships make traditional interpretation methods less effective. By offering a consistent and theoretically sound framework for model interpretation, SHAP enhances the transparency and trustworthiness of machine learning models, making it an essential tool in the era of explainable AI.

To calculate the SHAP value for a feature *i* in a prediction *f(x)*, all possible subsets of features are considered. The SHAP value measures the change in the model's prediction when the feature *i* is added to each subset. Mathematically, the SHAP value for feature *i* is given by the formula:

$$\phi_i(f) = \sum_{S \subseteq F \setminus \{i\}} \frac{|S|!(|F|-|S|-1)!}{|F|!} x[f(x_{S \cup \{i\}}) - f(x_S)]$$

SHAP values possess several desirable properties, including efficiency (the sum of the SHAP values for all features equals the difference between the prediction and the average prediction), symmetry (features contributing equally to all subsets receive the same SHAP value), dummy (features that do not change the prediction in any subset have a SHAP value of zero), and additivity (for models decomposable into additive components, the SHAP values of the components sum to the SHAP value of the model). These properties ensure that SHAP values provide a consistent and interpretable measure of feature importance, making them valuable for understanding complex models. SHAP is particularly useful in domains where model transparency is critical, such as healthcare, finance, and regulatory environments, ensuring that models are not only accurate but also interpretable and fair, thereby fostering trust and accountability in machine learning applications.

Figure 7 illustrates the SHAP values of the proposed model, revealing that the distance between the pile and the monitoring location exhibited the highest influence on peak particle velocity (PPV). An inverse relationship was observed between distance and PPV from pile driving, consistent with the principle of geometric damping where vibration energy dissipates as it propagates through the soil. The SHAP values demonstrated both positive and negative effects, as they are calculated relative to the model's average output. Values below the average resulted in positive Shapley values, augmenting the predicted model output. The second most influential factor was hammer weight, although its impact was significantly lower than that of distance. Higher hammer weights corresponded to increased energy transfer to the pile, resulting in greater vibration amplitude. This relationship aligns with the principle of energy conservation, where a larger input energy translates to higher ground vibrations. Pile size was the third most influential parameter, with larger piles generating more vibration due to Newton's Second Law of Motion (F = ma). Increased mass leads to higher ground acceleration, and larger piles also displace more soil during driving, contributing to greater vibration transmission. Sensor location emerged as the fourth most influential feature, with a notable difference observed between ground-based and building-mounted sensors. This variation can be attributed to the complex interaction between soil and structure, including potential amplification or attenuation effects in buildings. Pile length followed in order of influence, likely due to its role in determining the depth of soil interaction and the resulting vibration transmission pathways. Longer piles may reach stiffer soil layers, affecting vibration propagation characteristics. Drop height demonstrated a moderately low impact on PPV results, possibly because its effect is partially captured by the hammer weight feature. However, it still contributes to the overall energy input of the pile driving system.
The least influential feature was sensor orientation, indicating that differences in sensor alignment had a moderate effect on PPV monitoring outcomes. This suggests that while vibrations may have directional components, the peak values are relatively consistent across orientations, possibly due to the complex wave propagation in soil and structural systems. These findings highlight the multifaceted nature of pile driving vibrations and underscore the importance of considering multiple parameters in predictive models and vibration control strategies.

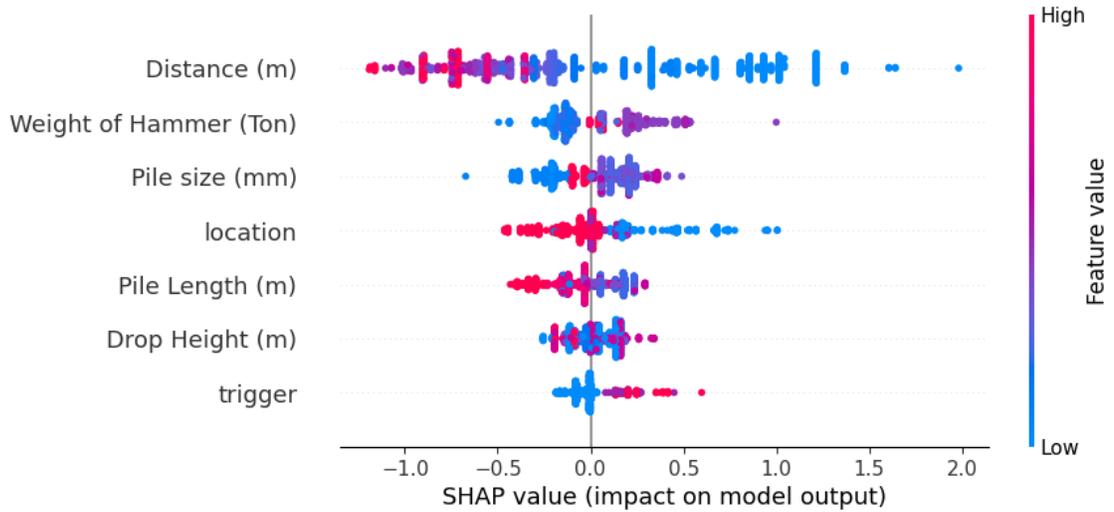

**Figure 7** The SHAP value of different feature of the model

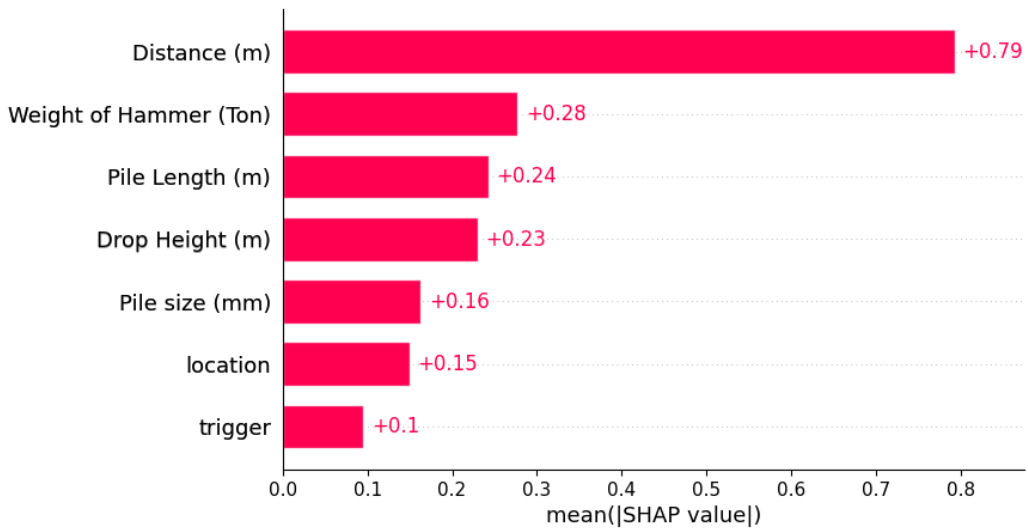

**Figure 8** The mean of SHAP value of different feature of the model

Figure 8 illustrates the relative impact of various features on Peak Particle Velocity (PPV) measurements. Quantitative analysis reveals a hierarchical influence, with distance exhibiting the most significant effect, followed by hammer weight, pile size, sensor location, pile length, drop height, and sensor arrangement, in descending order of importance. This hierarchy contrasts with the initial data analysis using mutual information (Fig. 3), where drop height was deemed less influential than pile size. Notably, SHAP (SHapley Additive exPlanations) values indicated a higher importance for drop height. Moreover, the relative importance of distance was substantially greater than other variables in the SHAP analysis, a distinction not observed in the mutual information approach. These findings underscore the efficacy of Explainable Artificial Intelligence (XAI) techniques in elucidating key factors affecting the phenomenon under investigation. Furthermore, this methodological approach facilitates the prediction and interpretation of model behavior trends, enhancing our understanding of PPV dynamics in pile driving operations.

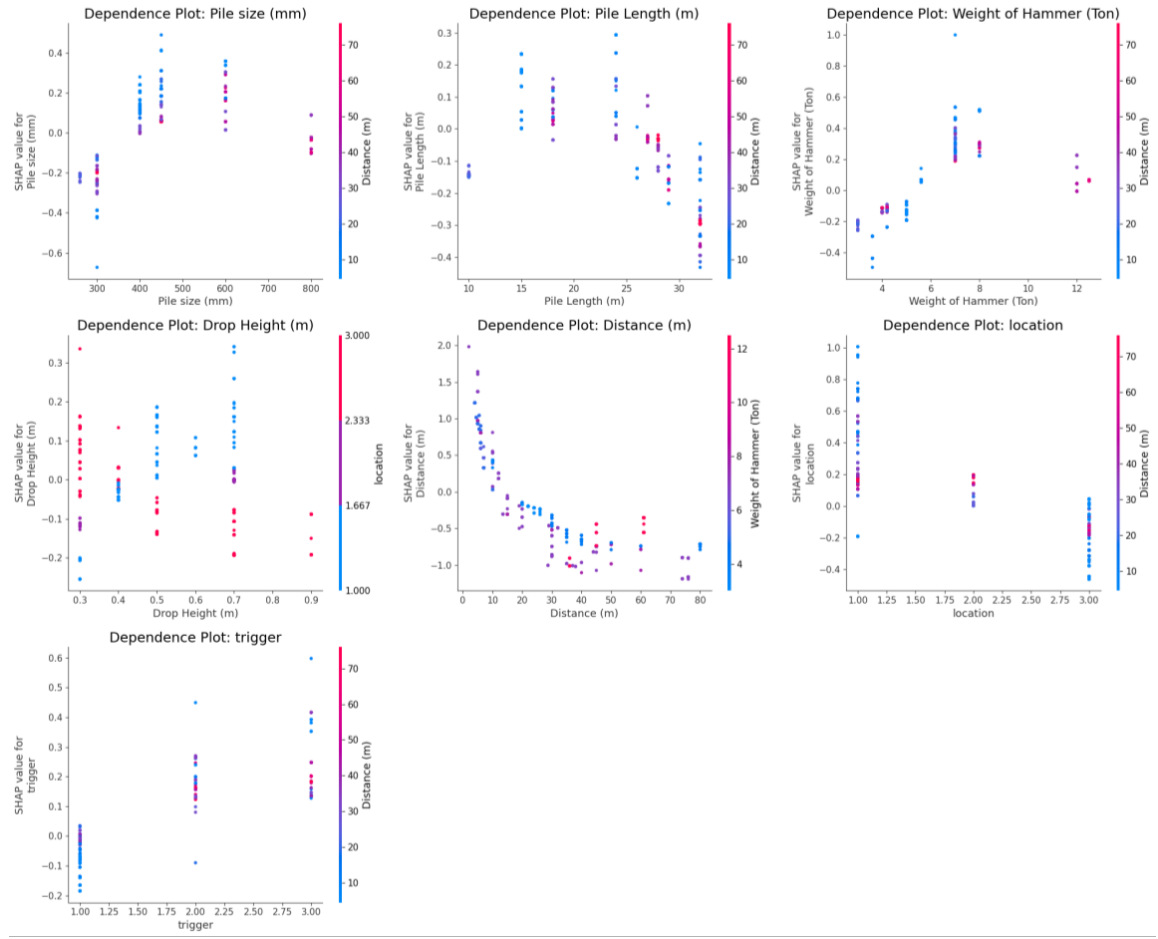

**Figure 9**  The comparison between feature in model and SHAP value

Figure 9 illustrates the complex relationships between SHAP (SHapley Additive exPlanations) parameters and various features influencing Peak Particle Velocity (PPV) in pile driving operations. This analysis reveals intricate interactions and potential threshold effects among the examined features. Distance emerges as the most influential factor, exhibiting an inverse correlation with SHAP values. This indicates that as distance from the pile driving location increases, the predicted PPV values decrease. Notably, the impact of distance on PPV diminishes significantly beyond 30 meters, suggesting a non-linear relationship between distance and vibration propagation. Hammer weight, identified as the second most critical feature, demonstrates a positive correlation with PPV values. However, a plateau or slight decrease in impact is observed for weights exceeding 8 tons. This phenomenon may indicate a saturation point, beyond which further increases in hammer weight yield diminishing returns in terms of PPV generation. This insight could have important implications for optimizing pile driving equipment selection. Pile size shows a positive impact on PPV up to approximately 600 mm, after which a slight reduction is noted. This non-monotonic relationship suggests that larger pile sizes may not always result in higher vibration levels, possibly due to changes in soil-pile interaction dynamics or energy dissipation mechanisms. Sensor location significantly influences PPV measurements, with ground-based sensors yielding the highest impact. This finding underscores the importance of consistent and strategic sensor placement in vibration monitoring protocols.

An inverse relationship is observed between pile length and PPV, possibly attributable to increased distance from the pile tip, which is hypothesized to be the primary source of vibration. Interestingly, shorter pile lengths (around 10 m) exhibit a lower influence compared to longer piles. This counterintuitive result warrants further investigation into the mechanics of vibration generation and propagation in piles of varying lengths. The impact of drop height on PPV demonstrates considerable variation, suggesting that changes in drop height do not consistently influence PPV values. This variability may be due to complex interactions with other parameters or site-specific conditions, highlighting the need for careful consideration of drop height in vibration prediction models. Regarding measurement direction, vertical sensor orientation appears to result in higher PPV values compared to longitudinal and lateral directions. This anisotropic behavior in vibration propagation could be attributed to the predominantly vertical force application during pile driving and the layered nature of most soil profiles.

This comprehensive analysis, leveraging SHAP values, provides nuanced insights into the relative importance and directional influence of key parameters on PPV predictions in pile driving operations. The observed complex interactions and potential threshold effects among the various features underscore the necessity of sophisticated, multi-parameter approaches in accurately predicting and mitigating pile driving-induced vibrations. These findings have significant implications for the optimization of pile driving practices, the development of more accurate predictive models, and the establishment of effective vibration control strategies in geotechnical engineering applications.

## 3.1 Comparison with the exiting approach

Figure 10 illustrates the comparison between field measurements and various pre-existing methods. The predictive models proposed by (Attewell and Farmer 1973) and (Achmus et al. 2010) appear to overestimate the results obtained from field tests. This discrepancy may be attributed to the damping properties of soft clay, which serve to attenuate the vibrations induced by pile driving. Despite its high compressibility and low shear strength, very soft clay can be effectively utilized as a damping medium to reduce vibration amplitudes (Teachavorasinskun et al. 2002; Garala and Madabhushi 2019). The substantial damping ratio inherent to soft clay layers can be harnessed to mitigate stress amplitudes, especially during seismic events. In contrast to sand, which generally exhibits lower damping characteristics, soft clay offers superior vibration isolation capabilities owing to its intrinsic dampness. Furthermore, the proposed model, which employs Artificial Neural Networks (ANN), demonstrates a more accurate alignment with the observed data. Notably, the Peak Particle Velocity (PPV) values at shorter distances from the pile driving site were observed to be higher than those predicted by empirical formulas.

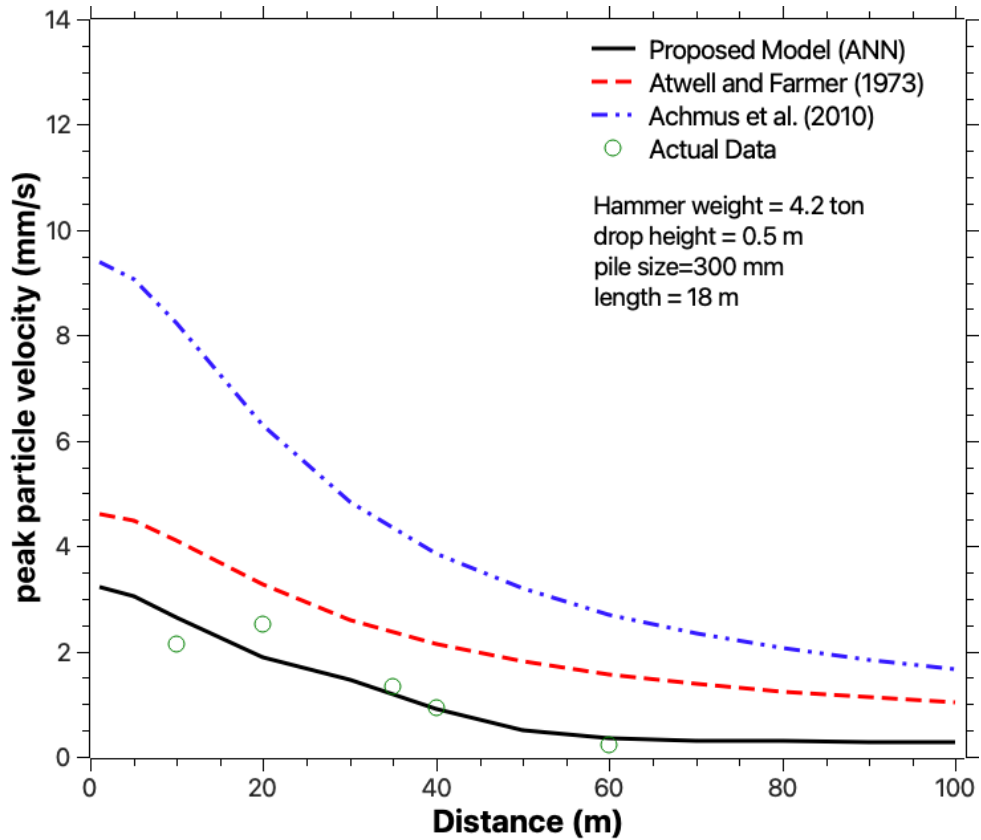

**Figure 10 Comparison of peak particle velocity from field with different predicting methods**

     A web-based application leveraging a trained machine learning model for peak particle velocity (PPV) prediction in pile driving operations has been developed and is accessible at https://piledriving.streamlit.appBuilt using the Streamlit framework, this application offers a user-friendly interface for interacting with complex algorithms. Users can input essential parameters including pile length, dimensions, hammer weight, drop height, sensor distance, location, and measurement direction. The application then processes these inputs through the trained model to generate an immediate PPV prediction, crucial for assessing potential vibration impacts on nearby structures and environments. This streamlined approach significantly reduces the time and complexity involved in traditional calculation methods, offering a convenient tool for both on-site and office-based assessments. The application's responsive design ensures cross-platform compatibility, enabling its use on various devices from smartphones to desktop computers. To promote transparency and reproducibility in geotechnical engineering

research, the complete source code and model weights are publicly available on GitHub (https://github.com/Sompote/pile_PPV). This development represents a significant advancement in applying machine learning to geotechnical challenges, bridging the gap between advanced predictive models and practical engineering applications in pile driving and vibration analysis.

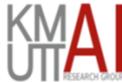

Figure 11 The web application for prediction of peak

## 4. Conclusion

This study presents a novel approach to predicting pile driving vibrations in Bangkok's soft clay subsoil using a deep neural network model. The developed model demonstrates superior accuracy compared to existing empirical formulas and other machine learning techniques, as evidenced by its lower Mean Absolute Error (MAE).

The application of SHAP (SHapley Additive exPlanations) analysis has revealed nuanced relationships between various parameters and their impacts on vibration prediction. Notably, distance emerged as the most influential factor, exhibiting an inverse correlation with peak particle velocity (PPV) and a significant reduction in impact beyond 30 meters. Hammer weight showed a positive correlation with PPV, but with diminishing returns beyond 8 tons. Pile size demonstrated a non-monotonic relationship, with peak influence around 600 mm. These insights provide a more comprehensive understanding of pile driving dynamics in soft clay conditions, challenging some conventional assumptions and highlighting the complex interplay of factors affecting vibration propagation. The SHAP analysis also revealed unexpected trends, such as the inverse relationship between pile length and PPV, and the significant influence of sensor location on measurements. These findings underscore the importance of considering multiple parameters in predictive models and vibration control strategies, moving beyond simplistic linear relationships often assumed in traditional approaches.

The development of a user-friendly web application based on this model bridges the gap between advanced machine learning techniques and practical engineering applications. By making the model and its source code publicly available, this research contributes to the transparency and reproducibility of geotechnical engineering studies, fostering further advancements in the field.

In conclusion, this study represents a significant step forward in the application of artificial intelligence to geotechnical challenges. The developed model and insights gained from the SHAP analysis have the potential to enhance the accuracy of vibration predictions, optimize pile driving practices, and improve vibration control strategies in urban construction projects. The non-linear relationships and threshold effects uncovered in this research open new avenues for investigation and may lead to more nuanced approaches in geotechnical engineering practice. Future research can build upon this foundation to further refine predictive models, explore additional applications of machine learning in geotechnical engineering, and investigate the underlying physical mechanisms behind the observed relationships.

**Declaration:**

- Funding: This research did not receive any specific grant from funding agencies in the public, commercial, or not-for-profit sectors.
- Conflicts of Interest: The authors declare no conflict of interest.
- Data Availability: The dataset used in this study is available upon reasonable request to the corresponding author.
- Code Availability: The deep neural network model developed in this study is publicly available on the authors' GitHub repository: https://github.com/Sompote/pile_PPV. The model can be implemented using Python.
- Software Availability: A web-based application for estimating pile driving vibrations based on this research has been developed and is freely accessible at: https://piledriving.streamlit.app.
- Author Contributions: Sompote Youwai: Conceptualization, Methodology, Software, Validation, Formal analysis, Investigation, Resources, Data Curation, Writing - Original Draft, Writing - Review & Editing, Visualization, Supervision, Project administration. Anuwat Pamungmoon: Methodology, Validation, Formal analysis, Data Curation.

## 5. Acknowledgement

## List of Symbol

| Symbol | Description |
|---|---|
| $y$ | Output tensor |
| $x$ | Input tensor |
| $A$ | Weight matrix |
| $b$ | bias |
| $z$ | Output from standard scaler |
| $x_i$ | *Data point* |
| $\mu$ | *Mean of data* |
| $\sigma$ | *Standard deviation* |
| $F$ | The set of all features |
| $x_S$ | The input features in subset S |
| $f(x_{S \cup \{i\}})$ | The model's prediction when feature *i* is included in the subset *S*. |
| $n$ | *The number of sample* |
| $\hat{y}_i$ | *The predicted output tensor* |
| $y_i$ | The actual output |
| $\phi_i(f)$ | The SHAP value |
| $S$ | *The all feature subset* |